\documentclass[journal]{IEEEtran}
\usepackage{graphicx}
\usepackage{amsfonts}

\usepackage{import}
\usepackage{transparent}

\usepackage[english]{babel}
\usepackage{amsmath, amssymb}
\usepackage{float}
\usepackage{makecell}
\usepackage{multirow}
\usepackage{cite}
\usepackage{amsmath,amssymb,amsfonts}
\usepackage{algorithmic}
\usepackage{graphicx}
\usepackage{textcomp}
\usepackage{xcolor}

\usepackage{subfigure}
\usepackage{epstopdf}

\usepackage{graphicx}
\usepackage{comment}
\usepackage{amsmath,amssymb} 
\usepackage{color}
\usepackage{enumerate}

\usepackage{booktabs}

\usepackage{soul}
\soulregister\cite7 
\soulregister\citep7 
\soulregister\citet7 
\soulregister\ref7 
\soulregister\pageref7 

\ifCLASSINFOpdf
\else
\fi
\begin{document}
%
\title{SC-NeRF: Self-Correcting Neural Radiance Field with Sparse Views}
%
%
%

\author{Liang Song, Guangming Wang,
       Jiuming Liu, Zhenyang Fu, Yanzi Miao, and Hesheng Wang 
        
\thanks{*This work was supported by the Fundamental Research
Funds for the Central Universities (Grant No.2020ZDPY0303), the General Program of National Natural Science Foundation of China (Grant No.61976218). The first two authors contributed equally. Corresponding Author: Yanzi Miao and Hesheng Wang.}
\thanks{L. Song, Z. Fu, and Y. Miao are with Engineering Research Center of Intelligent Control for Underground Space, Ministry of Education, School
of Information and Control Engineering, Advanced Robotics Research
Center, China University of Mining and Technology, Xuzhou 221116,
China.}
\thanks{G. Wang, J. Liu, and H. Wang are with Department of Automation, Key
Laboratory of System Control and Information Processing of Ministry of
Education, Key Laboratory of Marine Intelligent Equipment and System
of Ministry of Education, Shanghai Engineering Research Center of
Intelligent Control and Management, Shanghai Jiao Tong University,
Shanghai 200240, China.}
}

%
%

\markboth{Journal of \LaTeX\ Class Files,~Vol.~14, No.~8, August~2015}%
{Shell \MakeLowercase{\textit{et al.}}: Bare Demo of IEEEtran.cls for IEEE Journals}
%



\maketitle

\begin{abstract}
In recent studies, the generalization of neural radiance fields for novel view synthesis task has been widely explored. However, existing methods are limited to objects and indoor scenes. In this work, we extend the generalization task to outdoor scenes, trained only on object-level datasets. This approach presents two challenges. Firstly, the significant distributional shift between training and testing scenes leads to black artifacts in rendering results. Secondly, viewpoint changes in outdoor scenes cause ghosting or missing regions in rendered images. To address these challenges, we propose a geometric correction module and an appearance correction module based on multi-head attention mechanisms. We normalize rendered depth and combine it with light direction as query in the attention mechanism. Our network effectively corrects varying scene structures and geometric features in outdoor scenes, generalizing well from object-level to unseen outdoor scenes. 
Additionally, we use appearance correction module to correct appearance features, preventing rendering artifacts like blank borders and ghosting due to viewpoint changes. By combining these modules, our approach successfully tackles the challenges of outdoor scene generalization, producing high-quality rendering results. When evaluated on four datasets (Blender, DTU, LLFF, Spaces), our network outperforms previous methods. Notably, compared to MVSNeRF, our network improves average PSNR from 19.369 to 25.989, SSIM from 0.838 to 0.889, and reduces LPIPS from 0.265 to 0.224 on Spaces outdoor scenes.
\end{abstract}


%
\IEEEpeerreviewmaketitle

\begin{figure}[t]
		\vspace{-1mm}
	\centering
	\includegraphics[scale=0.45]{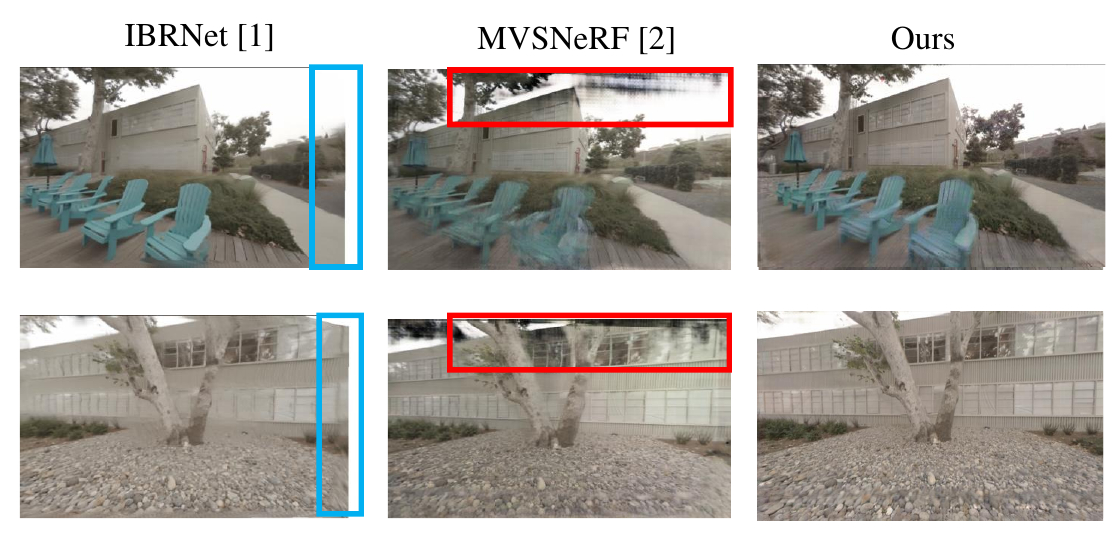}
	\vspace{-2mm}
	\caption{Comparison with previous methods IBRNet \cite{wang2021ibrnet} and MVSNeRF \cite{chen2021mvsnerf} on Spaces. We train both their and our networks on DTU and generalize to outdoor scenes in Spaces. The left, middle, and right images respectively show the rendering result of \cite{wang2021ibrnet}, \cite{chen2021mvsnerf}, and ours.}
	\label{fig:1}
\end{figure}

\section{Introduction}
%
%
%
%
\IEEEPARstart{N}{ovel } view synthesis (NVS) is a promising and long-standing problem that plays a fundamental role in  both the computer vision \cite{wang2021pwclo, 9042874}, robotic \cite{wang2020unsupervised} and graphics \cite{ngan2000visual}.

NVS aims to capture visual information from a sparse set of reference views to render an unseen target view. Early methods \cite{levoy1996light, chen1993view} produce a target view by interpolating in the ray \cite{levoy1996light} or pixel plane \cite{chen1993view}. Subsequent works \cite{buehler2001unstructured, bao2019depth} have exploited dense input views or geometric constraints, such as epipolar consistency \cite{buehler2001unstructured},  for depth-aware warping of the input views\cite{bao2019depth}. However, these methods are susceptible to artifacts caused by occlusion, the density of input views, and inaccurate geometry. To solve this problem, the multiplane image (MPI) approachs \cite{zhou2018stereo,flynn2019deepview} offer real-time rendering and generalization capabilities by representing the scene using a set of parallel planes derived from several input images. Nevertheless, when the perspective difference between the input view and target view is significant, there may be occurrences of edge rendering overlap \cite{zhou2023nerflix}. 

Recently, Neural radiance fields (NeRF) \cite{mildenhall2021nerf} and subsequent works \cite{liu2020neural,martin2021nerf} have the strong ability to produce realistic new view synthesis results. However, there are two main drawbacks: 1) It requires densely captured images for each scene. 2) It needs to be trained from scratch to overfit the new scene, with no generalization to unknown scenes.

To address the aforementioned shortcomings of NeRF, many methods \cite{trevithick2021grf,wang2021ibrnet,yu2021pixelnerf,liu2022neural} usually build a large composite dataset to fit the network to different scenarios, including object, indoor, and outdoor scenarios.  However, recent works \cite{chen2021mvsnerf,wang2021ibrnet} can not effectively generalize to outdoor scenes when trained on only object-level datasets. When MVSNeRF \cite{chen2021mvsnerf} generalizes to an outdoor scene, black artifacts appear in the sky or border, as shown in the blue box in Figure \ref{fig:1}. This is because the space scale and structure between the training scene and the test scene is extremely different and there maybe also exists reflective material in the outdoor scene. When the perspective gap between the input view and the target view increases, the result rendered by the IBRNet \cite{wang2021ibrnet} method will appear blank at the boundary, as shown in the red box in Figure \ref{fig:1}.

To solve these problems, we propose SC-NeRF, a novel approach that can be well generalized to different scenes by reconstructing radiation fields from only three unstructured multi-view input images. The SC-NeRF is trained only in an object-level dataset, while it can be generalized to a variety of different scenarios, especially outdoor scenarios. Due to the strong generalization ability, the SC-NeRF avoids time-consuming per-scene optimization and can directly regress realistic images from novel viewpoints of outdoor scenes.

To be specific, a low-resolution 3D geometric cost volume is constructed from sparse multi-view input images. This geometric cost volume can provide continuous geometric priors, when there is a non-covisual region  between the input and target perspectives. In order to solve the problem of artifacts in the rendered outdoor scene, the rendered features are corrected in terms of appearance and geometry. Specifically, a multi-head attention mechanism is leveraged to correct rendered characteristics using direction embedding as query, geometric or appearance features as key, and rendered features as value. Although it alleviates the shadow problem in the distance to some extent, it will cause shadow transfer in the render view. This is mainly because using only the direction as the query can not effectively get complete structure information of the scene. Therefore, we combine the rendered depth value with direction embedding as query, effectively solving the shadow transfer problem.

Our approach is completely differentiable, which can be trained in end-to-end manner from sparse view inputs. Our experiments show that with just three input views, our network can synthesize photo-realistic images on DTU \cite{jensen2014large}, Blender \cite{mildenhall2021nerf}, LLFF \cite{mildenhall2019local}, Spaces \cite{flynn2019deepview}. Overall, our contributions are as follows:
\begin{itemize}
\item  We propose a novel end-to-end network for synthesizing realistic images from sparse input views. We firstly propose a geometry correction module based on multi-head attention. It can address the issue of black artifacts in rendered views, caused by inconsistencies in scale and structure between training and testing scenes.
\item Building on the geometry correction module, we also design an appearance correction module to alleviate boundary blank and ghosting artifacts in rendered views caused by relatively large viewpoint changes.
\item We validate the effectiveness of our model on four datasets, including Blender, LLFF, DTU, and Spaces. Notably, on the outdoor scenes in the Spaces dataset, our model outperforms MVSNeRF by 34.17\% in terms of PSNR, and IBRNet by 19.9\%.
\end{itemize}

\section{Related Work}

\subsection{Novel View Synthesis via NeRF}
In recent years, various neural scene representations have been proposed to implement view synthesis \cite{ mildenhall2021nerf,thies2019deferred,lombardi2019neural,9466401,9320342}.
NeRF \cite{mildenhall2021nerf} has achieved very impressive results in novel view synthesis by optimizing the 5D neural radiation field of a scene. However, it must be optimized for each new scenario, which takes hours or days to converge.

There are some methods proposed to extend NeRF's generalization capabilities \cite{chen2021mvsnerf,trevithick2021grf,wang2021ibrnet,yu2021pixelnerf,chibane2021stereo}. GRF \cite{trevithick2021grf} projects the learned local image features onto three-dimensional points to obtain a general and rich point representation. MVSNeRF \cite{chen2021mvsnerf} utilizes plane sweep cost volume for neural radiation field reconstruction.  NeuRay \cite{liu2022neural} enables the construction of radiation fields to focus on visible image features by modeling the visibility of 3D points in the input view. However, none of these methods consider how to train the network only on an object-level dataset and be generalized to outdoor scenes.
\subsection{Multi-View Stereo}
Multi-view stereo (MVS) is a core problem in the field of computer vision. Multi-view stereo matching reconstruction can be regarded as the inverse process of taking pictures of a certain scene. Its purpose is to restore the real 3D scene through images taken from different viewpoints. A large number of traditional methods \cite{de1999poxels,esteban2004silhouette,furukawa2009accurate,schonberger2016pixelwise,seitz2006comparison} use hand-crafted similarity metrics and regularization methods to calculate dense correspondence of scenes. These methods can achieve good results on non-Lambertian surfaces and scenes without weakly textured regions. However, the artificially designed similarity metrics become unreliable in weakly textured regions, thus leading to incomplete reconstruction results.
Recently, deep learning techniques \cite{yao2018mvsnet,yao2019recurrent,chen2019point,gu2020cascade} have been introduced. Among these, MVSNet \cite{yao2018mvsnet} applies a 3D CNN for depth estimation on the plane scan cost of the reference view, and achieves high-quality 3D reconstruction. Subsequent works \cite{yao2019recurrent, chen2019point, gu2020cascade} extend this technique to recurrent planar sweeps \cite{yao2019recurrent}, point-based densification\cite{chen2019point}, and cascaded cost volume \cite{gu2020cascade} for improving the effect of reconstruction.
We follow their ideas to build a geometrically consistent cost volume. 
This ensures that the network meets the consistency of multiple views, so that the network can focus on information from different views and also learn geometric priors when rendering the novel views. 
\subsection{Transformer in NeRF}
Recently, there have been some attempts to incorporate the transformer \cite{vaswani2017attention} architecture into the NeRF model. IBRNet \cite{wang2021ibrnet} proposes ray transformers that dynamically correlate appearance information from multiple source views. NerFormer \cite{reizenstein2021common} proposes to use transformers to aggregate features between different views on the ray and learn radiance fields from the aggregated features. GNT \cite{wang2022attention} directly regresses colors to synthesize views without need for NeRF's volumetric rendering. GPNR \cite{suhail2022generalizable} improves generalization by using several stacked "patch-based" transformers to aggregate global features. Different from the above methods, our method mainly uses transformers to correct the features for better generalization to different scenes. Specifically, we use the transformer to correct the geometric and appearance features, which make better use of information from different perspectives and improve the generalization ability of network and geometric reconstruction.
\begin{figure*}[t]
	\centering
	\includegraphics[scale=0.6]{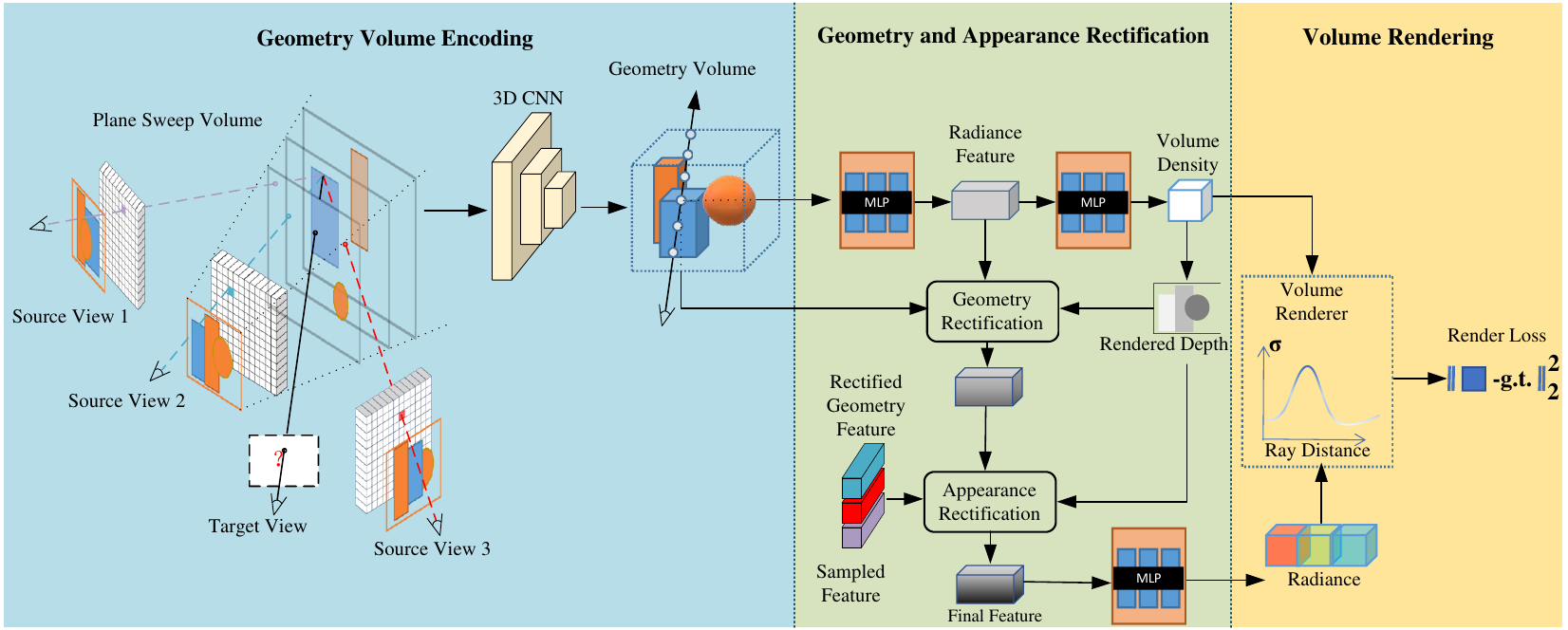}
	\caption{Overview of SC-NeRF. We first extract image features and warp them onto a plane sweep,then use 3DCNN to build a geometric Volume. Second, we use the geometric feature rectification and appearance feature rectification modules to obtain the final radiation features. Finally, we use an MLP to obtain the volume density and RGB radiation values of any sampling point in space, and use volume rendering to obtain the rendered view. }
	\label{fig:2}
\end{figure*}

\section{Method}
Given several sparse source views, our method uses volume rendering to synthesize a target view in a new camera pose. The core problem is how to obtain the density and colors of the continuous space by using the information from the input views, and how to make this representation generalize to other scenes, especially outdoor scenes.

The overview of our SC-NeRF is shown in Fig. \ref{fig:2}. For the sparse input $M$  views $(M=3)$, we first warp the extracted image features into the reference perspective and construct a geometric encoding volume using 3DCNN (Sec. \ref{3.1}). Then, we obtain the final radiance features through the geometric feature correction module and the appearance feature correction module based on the multi-head attention mechanism (Sec. \ref{3.2}). Finally, we use an multi-layer perceptron (MLP) to regress the volume density and RGB radiance from the corrected radiance features. These volume properties are passed through the volume rendering formula to obtain the final rendered images (Sec. \ref{3.3}).

\subsection{Geometry Volume Encoding} \label{3.1}
Inspired by the recent MVSNeRF \cite{chen2021mvsnerf}, we construct the encoding volume V at the reference view, allowing for geometry-aware scene understanding. 

First of all, a 2D CNN $G_{1}$ is used to extract the local appearance features of the input images. In our network, each input image $I_{i}\in  R^{H_{i}\times W_{i}\times 3 }$ is converted into a 2D feature map $F_{i}\in  R^{H_{i}/4 \times W_{i}/4\times C_{1}} $ by a down-sampled convolution operation:
\begin{equation}
F_{i}=G_{1}(I_{i} ),
\end{equation}
where $H_{i}$ and $W_{i}$ are the image height and width, and $C_{1}$ is
the number of image feature channels. 

Then, we transform the features of the source view into the reference view by the homographic warping operation. Given the camera intrinsic $\left [ K \right ]$ and extrinsic parameters $\left [ R,T \right  ]$, we use the homographic warping:
\begin{equation}
H_{i}(z) = K_{i} \cdot (R_{i} \cdot R_{r}^{T}+\frac{(t_{r}-t_{i}) \cdot n_{r} }{z}) \cdot K_{r}^{-1},
\end{equation}
where $H_{i}$ is the matrix warping from the view $i$ to the reference view $r$ at depth $z$. $K_{i}$ and $K_{r}$ are the intrinsic matrices. $n_{r}$ denotes the unit normal vector. $R$
and $t$ are the camera rotation and translation matrices. Each feature
map $F_{i}$ can be warped to the reference view by:
\begin{equation}
F_{i,z}(u,v)=Pad(F_{i})(H_{i}(z)\left [ u,v,1 \right ] ^{T}),
\end{equation}
where $F_{i,z}$ is the warped feature map at depth $z$, and $(u, v)$ represents a pixel location in the reference view. $Pad$ indicates the feature image edge padding operation. In this work, we parameterize (u, v, z) using the normalized device coordinate (NDC) at the reference view.

We leverage the variance-based method \cite{chen2021mvsnerf} to compute the cost from the warped feature maps on the $D$ sweeping planes. In particular, for each position $P$, its cost feature vector is computed by:
\begin{equation}
P(u,v,z) = Var(F_{i,z}(u,v)),
\end{equation}
where $Var$ is the variance operation.

Finally, we use a 3D CNN network $G_{2}$ with a U-Net structure to encode the cost volume mentioned above. This process is expressed by:
\begin{equation}
V = G_{2}(P),
\end{equation}
where $V$ is encoding geometry volume. This encoded volume contains the geometry feature of the scene, and is later continuously interpolated and converted into volume density.
\begin{figure*}[t]
	\centering
	\includegraphics[scale=0.4]{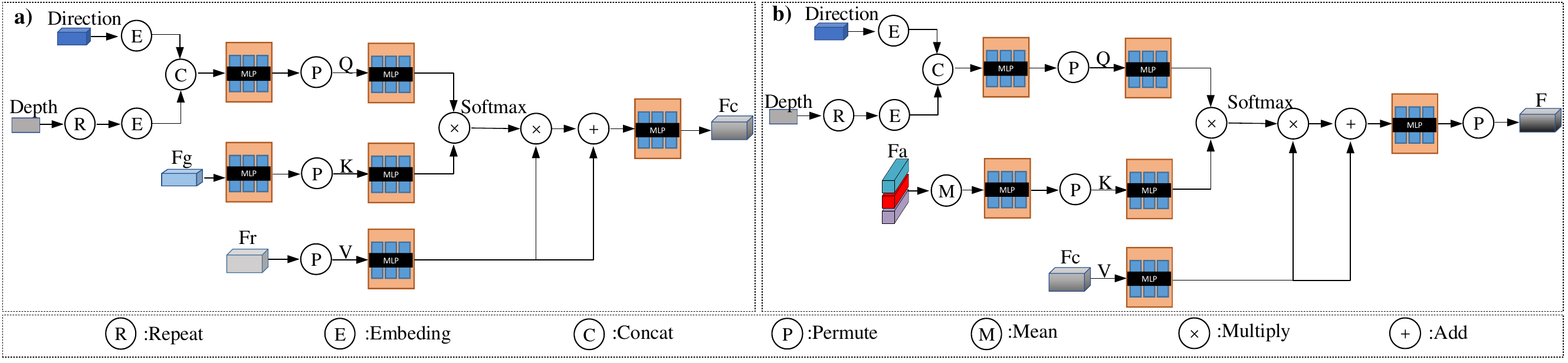}
	\caption{Geometry and appearance rectification modules. a) For the appearance features rectification module, direction embedding and depth embedding be used as query, sampled volume features $F_{g}$ are as key, the radiance features $F_{r}$ is taken as a value. b) For the appearance features rectification module, we also use the direction embedding and depth embedding as the query, the sampled image feature $F_{a}$ as the key, and the corrected feature $F_{c}$ as the value, so as to build a multi-head attention mechanism.}
	\label{net_2}
\end{figure*}
\subsection{Geometric and Appearance Features Rectification} \label{3.2}
Black artifacts appear in the rendering result of outdoor view, which are caused by the following reasons: 1) The spatial range of the object-level training set is much smaller than that of the outdoor test scenes. 2) There are some non-Lambertian reflective objects in the outdoor scene, which lead to the deviation of the feature. At the same time, when the viewpoint changes drastically, there may be significant differences between multiple viewpoints, resulting in incomplete coverage of details and textures on the object surface in a single viewpoint image. These uncovered details and textures can lead to visual artifacts, such as ghosting and boundary blanking. To overcome these challenges, we design geometric and appearance feature rectification modules based on multi-head attention mechanisms.
\subsubsection{Geometric feature Rectification}
Given an arbitrary 3D location $x$, an MLP $M_{1}$ be used to obtain radiance features $F_{r}$,
\begin{equation}
 F_{r} = M_{1}(E(x),s),
\end{equation}
where $s$ is the neural feature trilinearly interpolated from the volume $V$ at the location $x$. $E(.)$ indicates embedding operation.
Then, the corresponding volume density $\sigma$ is regressed by an MLP $M_{2}$,
\begin{equation}
 \sigma = M_{2}(F_{r}).
\end{equation}
We obtain the rendered depth $\hat{D}$ of the pixel corresponding to the $k$ sampled points based on the volume rendering formula:
\begin{equation}
\hat{D} =\sum_{k=1}^{N}T_{k}(1-exp(-\sigma_{k}))z_{k},
\end{equation}
\begin{equation}
T_{k}=exp(-\sum_{j=1}^{k-1}\sigma _{j}).
\end{equation}

Our geometric feature correction module is shown in Fig.\ref{net_2} (a). First, we normalize the rendered depth and get depth embedding. At the same time, we do the same operation with the ray direction. An MLP $M_{3}$ is used to process depth embedding and direction embedding to obtain query values $Q$:
\begin{equation} 
 Q = M_{3}(E(\hat{D}) \oplus E(d)),
\end{equation}
where $\oplus$ is concat operation, d is direction vector. It is worth noting that if we only use direction embedding as the query value, the rendered view will have black 
shadows and white holes. 

Then, the radiance features $F_{r}$ serve as the value $V$ and the volume features $s$ as the key $K$. The matching matrix of attention is calculated in the sampling point channel rather than the feature channel. The reason for doing this is to allow the model to independently learn which depth sample points contribute more to the rendering. This enables assigning higher weights to these depth sample points. By focusing on assigning weights to individual points rather than features for each sample point, we can better capture the informative points and improve the rendering quality. Therefore, we choose to pay attention in the sampling point dimension and calculate the attention weight for each sampling point to overcome the above problem.

Finally, we generate corrected features $F_{c}$ through multiple attention mechanisms:
\begin{equation} \label{ss}
F_{c} = Multihead(Q,K,V),
\end{equation}
\begin{equation}
Multihead(Q,K,V)=(head_{1} \oplus... \oplus head_{h})W^{O}
\end{equation}
\begin{equation}
head_{i} = Attention(QW_{i}^{Q},KW_{i}^{K},VW_{i}^{V}),
\end{equation}
\begin{equation}
Attention(Q,K,V) = softmax(\frac{QK^{T} }{\sqrt{d_{k} } } )V,
\end{equation}
where $QW_{i}^{Q}$ represents the weight matrix of the query ($Q$) for the i-th attention head, $KW_{i}^{K}$ represents the weight matrix of the key ($K$) for the i-th attention head, $VW_{i}^{V}$ represents the weight matrix of the value ($V$) for the i-th attention head. is the feature matrix, $\oplus$ is contact operation. This corrected feature can adapt to the change of scene space.

\begin{table*}[t]
	\begin{center}
		\caption{Quantitative results of novel view synthesis at synthetic and real data. We show averaged results of PSNRs, SSIMs and LPISs on three different datasets \cite{mildenhall2021nerf,yao2018mvsnet,mildenhall2019local}. We compare our method with concurrent NeRF-based methods \cite{yu2021pixelnerf,wang2021ibrnet,chen2021mvsnerf}.}
		\label{table:1}
		\begin{tabular}{cc|ccc|ccc|ccc}
                \hline
                &\multirow{2}{*}{Model}    &\multicolumn{3}{c|}{Synthetic Data(Blender \cite{mildenhall2021nerf})}  &\multicolumn{3}{c|}{Real Data(DTU \cite{jensen2014large})}
                &\multicolumn{3}{c}{LLFF \cite{mildenhall2019local}}\\
                & &PSNR↑ &SSIM↑ &LPIPS↓       &PSNR↑ &SSIM↑ &LPIPS↓ 
                &PSNR↑ &SSIM↑ &LPIPS↓\\
                \hline
                &PixelNeRF \cite{yu2021pixelnerf}        &7.39 &0.658 &0.411    &19.31 &0.789 &0.382          &11.24 &0.486 &0.671                  \\
                &IBRNet \cite{wang2021ibrnet}       &22.44 &0.874 &0.195  &26.04 &0.917 &0.190  &21.79 &0.786 &0.279 \\
                &MVSNeRF \cite{chen2021mvsnerf}  &23.62 &0.897 &0.176  &26.63 &0.931 &0.168  &21.93 &0.795 &\textbf{0.252}\\
                &Ours     & \textbf{23.753} &\textbf{0.90} &\textbf{0.174}  &\textbf{26.834}  &\textbf{0.934} &\textbf{0.167} &\textbf{22.087} &\textbf{0.80 } &0.253 \\
                \hline
		\end{tabular}
	\end{center}
\end{table*}

\begin{table*}[t]
	\begin{center}
		\caption{Quantitative results of novel view synthesis at outdoor data. We evaluated the PSNR, SSIMS, LPIPS indicators under different difficulty level settings of the Spaces dataset \cite{flynn2019deepview}. We compare our method with concurrent NeRF-based methods \cite{wang2021ibrnet,chen2021mvsnerf}.}
		\label{table:2}
		\begin{tabular}{cc|ccc|ccc|ccc}
                \hline
                &\multirow{2}{*}{Model}    &\multicolumn{3}{c|}{Small}  &\multicolumn{3}{c|}{Medium}     &\multicolumn{3}{c}{Large} \\
                & &PSNR↑ &SSIM↑ &LPIPS↓       &PSNR↑ &SSIM↑ &LPIPS↓  &PSNR↑ &SSIM↑ &LPIPS↓\\
                \hline
                &IBRNet \cite{wang2021ibrnet}       &21.667 &0.844 &0.297  &21.002 &0.824 &0.311 &20.346 &0.797 &0.349 \\
                &MVSNeRF \cite{chen2021mvsnerf}  &19.369 &0.838 &0.265  &19.196  &0.797 &0.310 &19.604  &0.772 &0.336\\
                &Ours     & \textbf{25.989} &\textbf{0.889} &\textbf{0.224}  &\textbf{24.962}  &\textbf{0.856} &\textbf{0.260}  &\textbf{24.584} &\textbf{0.843} &\textbf{0.272}\\
                \hline
		\end{tabular}
	\end{center}
\end{table*}

\subsubsection{Appearance feature Rectification}
Considering that large changes in viewpoint can lead to boundary blanking and artifacts in rendering, we also correct the appearance features through the multi-head attention mechanism. 

We reproject the pixel feature $F_{i}$ back onto the sample points along the light. We preliminarily regard pixel features as the appearance features of each position in the three-dimensional space along the direction of light. With this formula, each 3D point can theoretically have the corresponding 2D appearance feature. Specifically, given a three-dimensional point $x$, the observed 2D image $I_{i}$ with camera intrinsics $K$ and camera pose $\xi $ , the corresponding 2D appearance feature $F_{a}$ can be retrieved through the following reprojection operation:
\begin{equation}
F_{a} = \pi(\left \{ (F_{i} \oplus I_{i}),\xi,K  \right \},\left \{ x,y,z\right \} ),
\end{equation}
where the function $\pi(.)$ follows the principle of multi-view geometry \cite{andrew2001multiple}. If the point is inside the image, we simply select the nearest pixel using bilinear interpolation and index its features for the 3D point. If the point is outside the image, we assign a zero
vector to the 3D point, which means there is no information observed.

As in the geometric rectification section, we also use direction embedding and depth embedding as query values $Q$. In order to make better use of the appearance information of each source perspective, we calculate the mean of the appearance features of the three samples. 

We use the mean appearance feature as the key $K$ and the correction feature $F_{c}$ as the value $V$. We use formula (\ref{ss}) to obtain the final corrected radiance feature $F$. This process allows our network to use the input source view appearance feature to correct the appearance feature $F_{a}$ of the render view. $F$ can effectively adapt to both the appearance of the scene and the geometry of the scene.
Finally we use an MLP $M_{4}$ decoding structure similar to NeRF to get radiance value $c$ :
\begin{equation}
c = M_{4}(F,E(\hat{D}),E(d)).
\end{equation}

\subsection{Rendering and Training} \label{3.3}
The method described in the previous sections generates corrected radiance $c$ and density $\sigma$ values. To render the color $\hat{rgb}$ of ray through the scene, we first query the color and density of N samples on the ray, and then accumulate the color and density along them:
\begin{equation}
\hat{rgb} =\sum_{k=1}^{N}T_{k}(1-exp(-\sigma _{k}))c_{k},  
\end{equation}
\begin{equation}
T_{k}=exp(-\sum_{j=1}^{k-1}\sigma _{j}),
\end{equation}
where $\hat{rgb}$  is the final pixel color output, and $T_{k}$ represents the volume transmittance. 

This volume rendering is completely differentiable, so SC-NeRF can regress the final pixel color at the target view point from the sparse input views in an end-to-end manner. We use the $\ell_{2}$ norm of the rendered pixel versus the real pixel as a loss.

\begin{equation}
L = \left \| rgb -\hat{rgb}  \right \|_{2}^{2},
\end{equation}
where $rgb$ is the ground truth pixel color sampled from the target image $I_{t}$ at a novel viewpoint. $\left \| .\right \|_{2}^{2}$ denote the $\ell_{2}$ norm.

\section{Experiments} 
We evaluate our SC-NeRF on four datasets, namely DTU \cite{jensen2014large}, Blender\cite{martin2021nerf}, LLFF\cite{mildenhall2019local}, and Spaces \cite{flynn2019deepview}. In addition, we conduct the ablation studies of our network to demonstrate the effectiveness of each component. Extensive experiments show that our method exceeds the current state-of-the-art methods, especially when generalized to outdoor scenes.

\subsection{Experimental Settings}

\subsubsection{Dataset and evaluation settings}
\begin{figure}[t]
	\centering
	\includegraphics[scale=0.35]{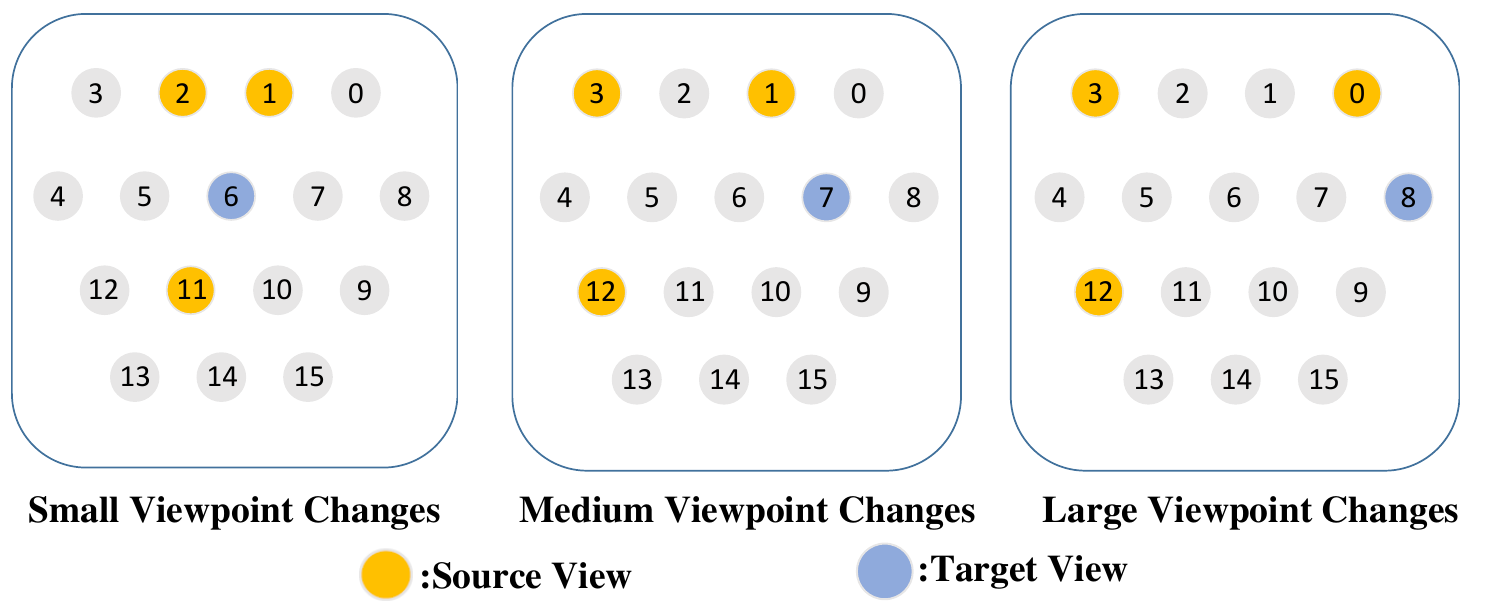}
	\caption{Three levels of viewpoint settings with increasing difficulty. We build three levels of difficulty rendering based on the gap between the source view and the target view. The orange ones represent the input 3 source views, and the blue ones represent the target view to be rendered. We call the different difficulty levels as "small", "medium" and "large" respectively.}
	\label{s}
\end{figure}

We train our network only on DTU dataset \cite{jensen2014large}. We follow the MVSNeRF \cite{chen2021mvsnerf} data partitioning method and divide the data into 88 training sences and 16 testing scenes. We also evaluate our method on additional synthetic \cite{martin2021nerf} and real datasets \cite{mildenhall2019local}, following MVSNeRF. In order to evaluate the generalization ability of the network to outdoor scenes, we also select corresponding scenes from the Spaces dataset \cite{flynn2019deepview} for evaluation. In particular, as shown in Figure \ref{s}, we selecte 8 outdoor scenes in the Spaces dataset which contains 100 scenes captured by a 16-camera rig, and set three different levels of difficulty through the spatial gap between the source and target views. The performance is evaluated by PSNR, SSIM \cite{wang2004image} and LPIPS \cite{zhang2018unreasonable} metrics.

\begin{figure*}[t]
	\centering
	\includegraphics[scale=0.8]{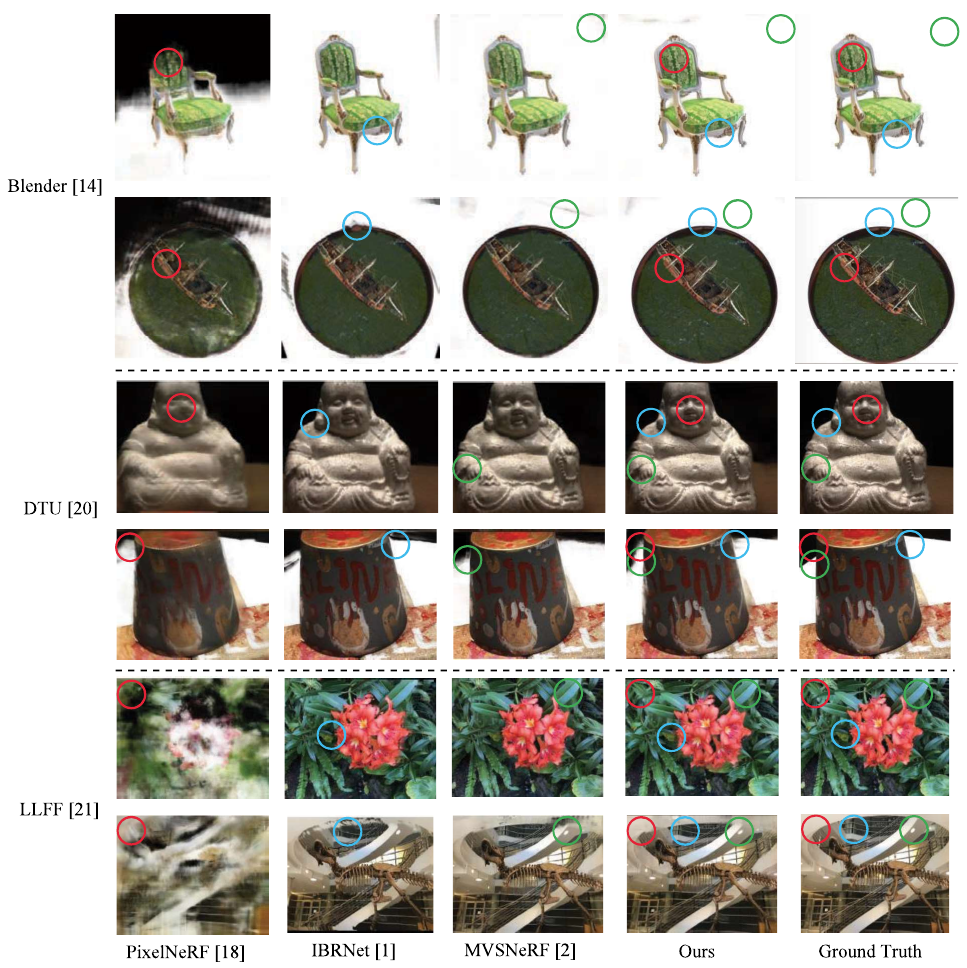}
	\caption{Rendering quality comparison at object level and indoor data. We show the visual comparison results of our method and other NeRF-based generalization methods \cite{yu2021pixelnerf,wang2021ibrnet,chen2021mvsnerf} on 3 different test sets \cite{mildenhall2021nerf,jensen2014large,mildenhall2019local}. For each data set, we select two sets of scenarios to show. From the red circle, it can be observed that PixelNeRF\cite{yu2021pixelnerf} has a poorer rendering effect. From the blue circle, it can be seen that IBRNet\cite{wang2021ibrnet} lacks sufficient detail in handling edge details. From the green circle, it can be noticed that MVSNeRF\cite{chen2021mvsnerf} is slightly inferior in rendering background details. }
	\label{img3}
\end{figure*}

\begin{table}[t]
	\begin{center}
		\caption{Quantitative results of novel view synthesis at outdoor data. We evaluated the PSNR, SSIMS, LPIPS indicators under different difficulty level settings of the DTU dataset \cite{jensen2014large}. We compare our method with concurrent NeRF-based methods \cite{yu2021pixelnerf,wang2021ibrnet,chen2021mvsnerf}.}
		\label{table:3}
		\begin{tabular}{ccccccc}
                \hline
                &Method &Abs err↓ &Acc (0.01)↑ &Acc (0.05)↑ \\
                \hline
                &PixelNeRF \cite{yu2021pixelnerf}  &0.239 &0.039  &0.187\\
                &IBRNet \cite{wang2021ibrnet} &1.62    &0.000      &0.001 \\
                &MVSNeRF \cite{chen2021mvsnerf} &0.035 &0.717      &0.866\\
                &Ours    &\textbf{0.022}  &\textbf{0.770}   &\textbf{0.914}                                      \\
                \hline
		\end{tabular}
	\end{center}
\end{table}
\subsubsection{Implement Details}
The dimension of image features $F_{i}$ extracted by 2D CNN is set as 32. The depth sampling planes for the homographic warping operation are 128. The volume feature $V$ dimension of 3D CNN encoding is set as 8. The number of sampling points for each ray is 128. The dimensions of position embedding, orientation embedding $E(\hat{d})$ and depth embedding $E(\hat{D})$ are 63, 33 and 11, respectively. The number of heads of multi-head attention used for appearance correction and geometry correction is 4. All training and evaluation experiments are conducted on a single RTX 3090 GPU with PyTorch1.10.1. We randomly select 1024 pixels from a novel viewpoint as a batch and apply the Adam optimizer with an initial learning rate of 0.0005.

\begin{figure*}[t]
	\centering
	\includegraphics[scale=0.85]{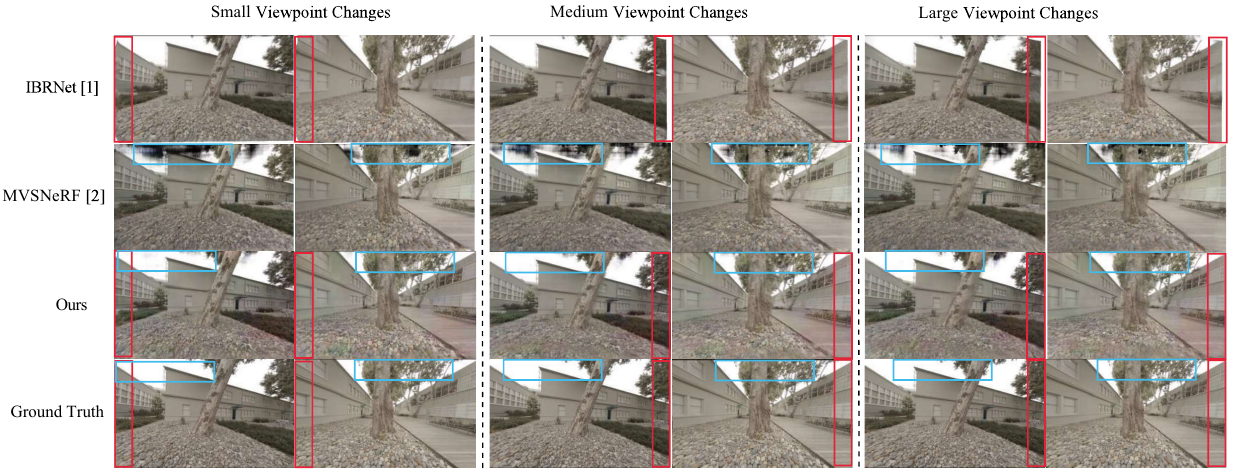}
	\caption{Rendering quality comparison at outdoor data. We show the results of a visual comparison of our method with two state-of-the-art methods \cite{chen2021mvsnerf,wang2021ibrnet} on the Space dataset \cite{flynn2019deepview} for three settings of different difficulty levels. From the red box, it can be observed that as the viewing angle increases, IBRNet \cite{wang2021ibrnet} produces blank spaces at the edges of the rendered image. From the blue box, it can be seen that MVSNeRF \cite{chen2021mvsnerf} exhibits black pseudo-shadows in areas such as the sky. Our method can effectively solve these issues in outdoor scenes. }
	\label{img_6}
\end{figure*}

\begin{table*}[t]
	\begin{center}
		\caption{Qualitative ablation study. We evaluated the PSNR, SSIMS, LPIPS indicators under different difficulty level settings of the Spaces dataset \cite{flynn2019deepview}. We compare the performance of the different components of our method.}
		\label{table:4}
		\begin{tabular}{cc|ccc|ccc|cccc}
               \hline
                &\multirow{2}{*}{Model}    &\multicolumn{3}{c|}{Small}  &\multicolumn{3}{c|}{Medium}     &\multicolumn{3}{c}{Large} \\
                & &PSNR↑ &SSIM↑ &LPIPS↓       &PSNR↑ &SSIM↑ &LPIPS↓  &PSNR↑ &SSIM↑ &LPIPS↓\\
               \hline
                &Baseline(BL)      &19.369 &0.838 &0.265  &19.196  &0.797 &0.310 &19.604  &0.772 &0.336\\                                         
               \hline
                &BL+Appearance-V(A\_V)     &23.037 &0.806 &0.393  &22.9 &0.796 &0.400 &23.107 &0.869 &0.439                                             \\
                &BL+Appearance-VD(A\_VD)    &23.894 &0.836 &0.310 &23.338 &0.808 &0.332 &23.148 &0.798 &0.341                                          \\
               \hline
                &BL+Geometry-V(G\_V)  &19.175 &0.822 &0.335 &18.706 &0.799 &0.359 &18.582 &0.787 &0.370                                                      \\
                &BL+Geometry-VD(G\_VD) &25.528 &0.880 &0.241 &24.220 &0.825 &0.298 &23.384 &0.810 &0.317                                                          \\
               \hline
                &BL+Appearance+Geometry(AG\_VD) &24.145 &0.845 &0.294 &23.374 &0.789 &0.335 &22.469 &0.772 &0.353 \\
                &BL+Geometry+Appearance(GA\_VD)    & \textbf{25.989} &\textbf{0.889} &\textbf{0.224}  &\textbf{24.962}  &\textbf{0.856} &\textbf{0.260}  &\textbf{24.584} &\textbf{0.843} &\textbf{0.272}\\
               \hline
		\end{tabular}
	\end{center}
\end{table*}

\subsection{Comparison Results}
We compare with three recent NeRF-based works, PixelNeRF \cite{yu2021pixelnerf}, IBRNet\cite{wang2021ibrnet}, and MVSNeRF \cite{chen2021mvsnerf} that also aim to improve generalization ability of NeRF. We input the three source views to retrain the three models on the DTU data for the fair comparison. We choose four groups of images in each scene from the three datasets \cite{mildenhall2021nerf,yao2018mvsnet,mildenhall2019local} for testing, and finally evaluate the performance with the mean PSNR, SSIM, and LPIPS. We show the quantitative results in Tab. \ref{table:1}. To further compare the generalization capability of our network in outdoor environments, we compare ours with the MVSNeRF and IBRNet in Tab. \ref{table:2}. For a more intuitive comparison of experimental effects, we show visualization comparison in Fig. \ref{img3} and Fig. \ref{img_6}. We also present the comparative results of the depth estimates at DTU dataset in Tab. \ref{table:3}.

\subsubsection{Comparisons of view synthesis at synthetic and indoor data}
Quantitative results in Tab. \ref{table:1} show that our SC-NeRF
performs the best in all datasets. Although our model is only trained on DTU, it can be well generalized to other two datasets with highly different distributions of scenes and views. On the Blender, DTU, and LLFF datasets, the PSNR evaluation values are higher than those of PixelNeRF by 16.363, 7.524, and 10.847 respectively. They are also higher than IBRNet by 1.313, 0.794, and 0.297 respectively, and higher than MVSNeRF by 0.133, 0.204, and 0.157 respectively. As shown in Fig. \ref{img3}, PixelNeRF has obvious blurring and artifacts when generalized to other scenes. This is because they only consider introducing 2D image features into the NeRF model, but don't consider the scene geometry. IBRNet has achieved well generalization results due to the introduction of rays transformer, but some artifacts still appear in the details. The view rendered by MVSNeRF tends to contain artifacts around the background because its cost volume is built for a specific reference view where the camera setback may not cover the target view sufficiently. The main reason for the superiority of our model is that we not only consider the geometric features of the scene, but also its appearance features, and use the rectification-based strategy to make the two features mutually optimized. This correction mechanism can improve the rendering effect in indoor scenes for better generalization.

\begin{figure*}[t]
	\centering
	\includegraphics[scale=0.6]{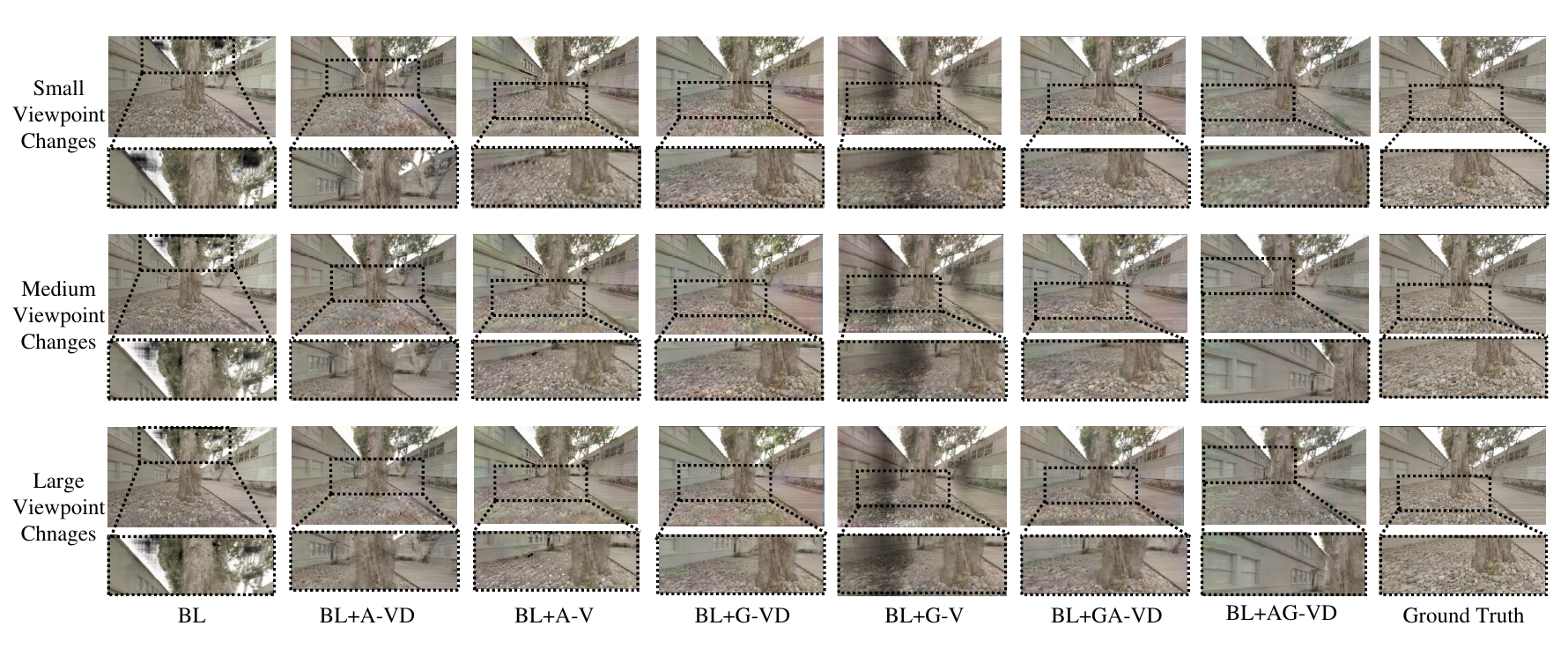}
	\caption{Qualitative ablation study. We show the visualization results of adding geometry correction module and appearance correction module on Baseline. We mark the area of our concern with a box to highlight problems such as rendering holes and artifact transfer.}
	\label{img_5}
\end{figure*}

\subsubsection{Comparisons of view synthesis at outdoor data}
To evaluate the ability to generalize to outdoor scenes, we test IBRNet, MVSNeRF, and our model on the Spaces dataset with three difficulty level settings. From Tab. \ref{table:2}, it can be clearly found that the generalization to outdoor scenes has a significant decline in the performance of their method compared with the test scene in the object level and indoor scenes. This can be seen in the degradation of their method's performance when generalized to outdoor scenarios. However, our method can achieve the best performance. On the three difficulty levels (small, medium, large) of the Spaces dataset, the PSNR obtained from testing is higher than IBRNet by 20\%, 18.9\%, and 20.1\% respectively. It is also higher than MVSNeRF by 34.2\%, 30\%, and 25.4\% respectively. From Fig. \ref{img_6}, we can analyze the reasons for the bad generalization of their two methods to outdoor scenes. Due to the increasing gap between the input view of the scene and the rendered target view, it can be found that IBRNet can not render the non-common view regions. Therefore, the rendered view will have a rendering blank area in the border outline. From the second line of Fig. \ref{img_6}, it can be found that MVSNeRF will have artifacts in the sky. This is because of a huge difference in scene depth between the training and testing set. Differently, we can effectively alleviate the depth inconsistency between the training and testing set by normalizing the rendered depth and embedding it as a part of the query value. At the same time, we use the appearance correction strategy to effectively use the appearance characteristics of different views to overcome the problem of rendering loss in non-common view areas. It can be seen from both qualitative results and quantitative metrics that our model outperforms the two state-of-the-art methods on the above comparison in outdoor scene rendering performance.

\subsection{Comparisons of Depth Reconstruction}

In order to evaluate whether the model effectively learns the ability to model the geometry of the scene. We reconstruct the depth as \cite{martin2021nerf} by weighting the depth values of the sampling points on the ray and the volume density. We compare our approach with three NeRF-based methods \cite{chen2021mvsnerf,wang2021ibrnet,yu2021pixelnerf}. It can be seen from the Tab. \ref{table:3} that our method achieves the best results of the estimation of the depth of the new view. it can be observed that the absolute error obtained from the tests is reduced by 90.8\% compared to PixelNeRF, by 98\% compared to IBRNet, and by 37\% compared to MVSNeRF. Since only the local features of the image are used and the geometric structure of the scene is not considered, the rendering depth of PixelNeRF has 20 times larger errors. It is worth noting that although IBRNet can render the target view well and has strong generalization ability, it does not learn the 3D model of the scene in essence, but only an interpolation synthesis in appearance. So it suffers from extremely poor depth estimations. Due to the geometric correction strategy adopted by our method, our method can outperform MVSNeRF's depth estimation metrics. 

\subsection{Ablations and Analysis}
Tab. \ref{table:4} and Fig. \ref{img_5} summarize the quantitative and qualitative results of SC-NeRF for different architecture choices at different difficulty levels on the Spaces dataset. We take MVSNeRF as our baseline. We will only add the appearance correction module with orientation embedding as query value on the baseline called "Appearanc-V". "Appearance-VD" means adding an appearance correction module with orientation embedding and rendering depth embedding as query values. Similarly, for the geometry correction module, we distinguish them by different query values, namely "Geometry-V" and "Geometry-VD".

From Tab. \ref{table:4}, we find that when only using direction embedding as the query, the geometric rectification  will make the indicators decrease, while the appearance rectification can effectively improve the PSNR. This shows that direction embedding as a query can effectively correct appearance characteristics. When we combine depth embedding with direction embedding as a multi-head attention query, we find that it can greatly improve the performance of geometric rectification, and also promote the appearance rectification. This shows that providing geometric depth information of the scene can effectively improve the performance of novel view rendering.

It can be seen from Fig. \ref{img_5}, when only the direction is used as the query value, white holes will appear in appearance rectification, and black shadow transfer will appear in geometric rectification. However, when the depth embedding is also used as the query value, it can solve the two problems of the above. 

Only using the appearance rectification module can improve the PSNR metrics very well, but the structural index SSIM will be lower than the Baseline. On the contrary, if only the geometric correction module is used, the SSIM and LPIPS indicators can be improved very well. 

By comparing the sequence of geometric correction and appearance correction, we can find that the feature planning and correction before the appearance correction of the feature can obtain better rendering performance. Therefore, our final model firstly corrects geometric features, and then appearance features.  

\section{CONCLUSIONS}
We propose a new generalizable approach for neural rendering. It provides a more practical neural rendering technique using a small number of images as input. Through the proposed geometric feature correction module and appearance feature correction module, our network can be trained on only object-level scenes to effectively generalize to outdoor scenes. We show that our rectification strategy can provide valuable geometry and appearance cues, leading to state-of-the-art performance under several challenging settings on four benchmark datasets.


%

\ifCLASSOPTIONcaptionsoff
  \newpage
\fi

\bibliographystyle{IEEEtran}  
\bibliography{IEEEabrv,bare_jrnl} 

\begin{thebibliography}{10}
\providecommand{\url}[1]{#1}
\csname url@samestyle\endcsname
\providecommand{\newblock}{\relax}
\providecommand{\bibinfo}[2]{#2}
\providecommand{\BIBentrySTDinterwordspacing}{\spaceskip=0pt\relax}
\providecommand{\BIBentryALTinterwordstretchfactor}{4}
\providecommand{\BIBentryALTinterwordspacing}{\spaceskip=\fontdimen2\font plus
\BIBentryALTinterwordstretchfactor\fontdimen3\font minus
  \fontdimen4\font\relax}
\providecommand{\BIBforeignlanguage}[2]{{%
\expandafter\ifx\csname l@#1\endcsname\relax
\typeout{** WARNING: IEEEtran.bst: No hyphenation pattern has been}%
\typeout{** loaded for the language `#1'. Using the pattern for}%
\typeout{** the default language instead.}%
\else
\language=\csname l@#1\endcsname
\fi
#2}}
\providecommand{\BIBdecl}{\relax}
\BIBdecl

\bibitem{ngan2000visual}
K.~N. Ngan, T.~Sikora, and M.-T. Sun, ``Visual communications and image
  processing 2000,'' \emph{Visual Communications and Image Processing 2000},
  vol. 4067, 2000.

\bibitem{levoy1996light}
M.~Levoy and P.~Hanrahan, ``Light field rendering,'' in \emph{Proceedings of
  the 23rd annual conference on Computer graphics and interactive techniques},
  1996, pp. 31--42.

\bibitem{chen1993view}
S.~Chen and L.~Williams, ``View interpolation for image synthesis proceedings
  of the 20th annual conference on computer graphics and interactive
  techniques,'' 1993.

\bibitem{buehler2001unstructured}
C.~Buehler, M.~Bosse, L.~McMillan, S.~Gortler, and M.~Cohen, ``Unstructured
  lumigraph rendering,'' in \emph{Proceedings of the 28th annual conference on
  Computer graphics and interactive techniques}, 2001, pp. 425--432.

\bibitem{zhou2018stereo}
T.~Zhou, R.~Tucker, J.~Flynn, G.~Fyffe, and N.~Snavely, ``Stereo magnification:
  Learning view synthesis using multiplane images,'' in \emph{SIGGRAPH}, 2018.

\bibitem{flynn2019deepview}
J.~Flynn, M.~Broxton, P.~Debevec, M.~DuVall, G.~Fyffe, R.~Overbeck, N.~Snavely,
  and R.~Tucker, ``Deepview: View synthesis with learned gradient descent,'' in
  \emph{Proceedings of the IEEE/CVF Conference on Computer Vision and Pattern
  Recognition}, 2019, pp. 2367--2376.

\bibitem{mildenhall2021nerf}
B.~Mildenhall, P.~P. Srinivasan, M.~Tancik, J.~T. Barron, R.~Ramamoorthi, and
  R.~Ng, ``Nerf: Representing scenes as neural radiance fields for view
  synthesis,'' \emph{Communications of the ACM}, vol.~65, no.~1, pp. 99--106,
  2021.

\bibitem{liu2020neural}
L.~Liu, J.~Gu, K.~Zaw~Lin, T.-S. Chua, and C.~Theobalt, ``Neural sparse voxel
  fields,'' \emph{Advances in Neural Information Processing Systems}, vol.~33,
  pp. 15\,651--15\,663, 2020.

\bibitem{martin2021nerf}
R.~Martin-Brualla, N.~Radwan, M.~S. Sajjadi, J.~T. Barron, A.~Dosovitskiy, and
  D.~Duckworth, ``Nerf in the wild: Neural radiance fields for unconstrained
  photo collections,'' in \emph{Proceedings of the IEEE/CVF Conference on
  Computer Vision and Pattern Recognition}, 2021, pp. 7210--7219.

\bibitem{chen2021mvsnerf}
A.~Chen, Z.~Xu, F.~Zhao, X.~Zhang, F.~Xiang, J.~Yu, and H.~Su, ``Mvsnerf: Fast
  generalizable radiance field reconstruction from multi-view stereo,'' in
  \emph{Proceedings of the IEEE/CVF International Conference on Computer
  Vision}, 2021, pp. 14\,124--14\,133.

\bibitem{trevithick2021grf}
A.~Trevithick and B.~Yang, ``Grf: Learning a general radiance field for 3d
  representation and rendering,'' in \emph{Proceedings of the IEEE/CVF
  International Conference on Computer Vision}, 2021, pp. 15\,182--15\,192.

\bibitem{wang2021ibrnet}
Q.~Wang, Z.~Wang, K.~Genova, P.~P. Srinivasan, H.~Zhou, J.~T. Barron,
  R.~Martin-Brualla, N.~Snavely, and T.~Funkhouser, ``Ibrnet: Learning
  multi-view image-based rendering,'' in \emph{Proceedings of the IEEE/CVF
  Conference on Computer Vision and Pattern Recognition}, 2021, pp. 4690--4699.

\bibitem{yu2021pixelnerf}
A.~Yu, V.~Ye, M.~Tancik, and A.~Kanazawa, ``pixelnerf: Neural radiance fields
  from one or few images,'' in \emph{Proceedings of the IEEE/CVF Conference on
  Computer Vision and Pattern Recognition}, 2021, pp. 4578--4587.

\bibitem{liu2022neural}
Y.~Liu, S.~Peng, L.~Liu, Q.~Wang, P.~Wang, C.~Theobalt, X.~Zhou, and W.~Wang,
  ``Neural rays for occlusion-aware image-based rendering,'' in
  \emph{Proceedings of the IEEE/CVF Conference on Computer Vision and Pattern
  Recognition}, 2022, pp. 7824--7833.

\bibitem{jensen2014large}
R.~Jensen, A.~Dahl, G.~Vogiatzis, E.~Tola, and H.~Aan{\ae}s, ``Large scale
  multi-view stereopsis evaluation,'' in \emph{Proceedings of the IEEE
  conference on computer vision and pattern recognition}, 2014, pp. 406--413.

\bibitem{mildenhall2019local}
B.~Mildenhall, P.~P. Srinivasan, R.~Ortiz-Cayon, N.~K. Kalantari,
  R.~Ramamoorthi, R.~Ng, and A.~Kar, ``Local light field fusion: Practical view
  synthesis with prescriptive sampling guidelines,'' \emph{ACM Transactions on
  Graphics (TOG)}, vol.~38, no.~4, pp. 1--14, 2019.

\bibitem{thies2019deferred}
J.~Thies, M.~Zollh{\"o}fer, and M.~Nie{\ss}ner, ``Deferred neural rendering:
  Image synthesis using neural textures,'' \emph{Acm Transactions on Graphics
  (TOG)}, vol.~38, no.~4, pp. 1--12, 2019.

\bibitem{lombardi2019neural}
\BIBentryALTinterwordspacing
S.~Lombardi, T.~Simon, J.~Saragih, G.~Schwartz, A.~Lehrmann, and Y.~Sheikh,
  ``Neural volumes: Learning dynamic renderable volumes from images,''
  \emph{ACM Trans. Graph.}, vol.~38, no.~4, jul 2019. [Online]. Available:
  \url{https://doi.org/10.1145/3306346.3323020}
\BIBentrySTDinterwordspacing

\bibitem{chibane2021stereo}
J.~Chibane, A.~Bansal, V.~Lazova, and G.~Pons-Moll, ``Stereo radiance fields
  (srf): Learning view synthesis for sparse views of novel scenes,'' in
  \emph{Proceedings of the IEEE/CVF Conference on Computer Vision and Pattern
  Recognition}, 2021, pp. 7911--7920.

\bibitem{de1999poxels}
J.~S. De~Bonet and P.~Viola, ``Poxels: Probabilistic voxelized volume
  reconstruction,'' in \emph{Proceedings of International Conference on
  Computer Vision (ICCV)}, vol.~2, 1999, p.~3.

\bibitem{esteban2004silhouette}
C.~H. Esteban and F.~Schmitt, ``Silhouette and stereo fusion for 3d object
  modeling,'' \emph{Computer Vision and Image Understanding}, vol.~96, no.~3,
  pp. 367--392, 2004.

\bibitem{furukawa2009accurate}
Y.~Furukawa and J.~Ponce, ``Accurate, dense, and robust multiview stereopsis,''
  \emph{IEEE transactions on pattern analysis and machine intelligence},
  vol.~32, no.~8, pp. 1362--1376, 2009.

\bibitem{schonberger2016pixelwise}
J.~L. Sch{\"o}nberger, E.~Zheng, J.-M. Frahm, and M.~Pollefeys, ``Pixelwise
  view selection for unstructured multi-view stereo,'' in \emph{Computer
  Vision--ECCV 2016: 14th European Conference, Amsterdam, The Netherlands,
  October 11-14, 2016, Proceedings, Part III 14}.\hskip 1em plus 0.5em minus
  0.4em\relax Springer, 2016, pp. 501--518.

\bibitem{seitz2006comparison}
S.~M. Seitz, B.~Curless, J.~Diebel, D.~Scharstein, and R.~Szeliski, ``A
  comparison and evaluation of multi-view stereo reconstruction algorithms,''
  in \emph{2006 IEEE computer society conference on computer vision and pattern
  recognition (CVPR'06)}, vol.~1.\hskip 1em plus 0.5em minus 0.4em\relax IEEE,
  2006, pp. 519--528.

\bibitem{yao2018mvsnet}
Y.~Yao, Z.~Luo, S.~Li, T.~Fang, and L.~Quan, ``Mvsnet: Depth inference for
  unstructured multi-view stereo,'' in \emph{Proceedings of the European
  conference on computer vision (ECCV)}, 2018, pp. 767--783.

\bibitem{yao2019recurrent}
Y.~Yao, Z.~Luo, S.~Li, T.~Shen, T.~Fang, and L.~Quan, ``Recurrent mvsnet for
  high-resolution multi-view stereo depth inference,'' in \emph{Proceedings of
  the IEEE/CVF conference on computer vision and pattern recognition}, 2019,
  pp. 5525--5534.

\bibitem{chen2019point}
R.~Chen, S.~Han, J.~Xu, and H.~Su, ``Point-based multi-view stereo network,''
  in \emph{Proceedings of the IEEE/CVF international conference on computer
  vision}, 2019, pp. 1538--1547.

\bibitem{gu2020cascade}
X.~Gu, Z.~Fan, S.~Zhu, Z.~Dai, F.~Tan, and P.~Tan, ``Cascade cost volume for
  high-resolution multi-view stereo and stereo matching,'' in \emph{Proceedings
  of the IEEE/CVF conference on computer vision and pattern recognition}, 2020,
  pp. 2495--2504.

\bibitem{vaswani2017attention}
A.~Vaswani, N.~Shazeer, N.~Parmar, J.~Uszkoreit, L.~Jones, A.~N. Gomez,
  {\L}.~Kaiser, and I.~Polosukhin, ``Attention is all you need,''
  \emph{Advances in neural information processing systems}, vol.~30, 2017.

\bibitem{reizenstein2021common}
J.~Reizenstein, R.~Shapovalov, P.~Henzler, L.~Sbordone, P.~Labatut, and
  D.~Novotny, ``Common objects in 3d: Large-scale learning and evaluation of
  real-life 3d category reconstruction,'' in \emph{Proceedings of the IEEE/CVF
  International Conference on Computer Vision}, 2021, pp. 10\,901--10\,911.

\bibitem{wang2022attention}
\BIBentryALTinterwordspacing
M.~V. T, P.~Wang, X.~Chen, T.~Chen, S.~Venugopalan, and Z.~Wang, ``Is attention
  all that ne{RF} needs?'' in \emph{The Eleventh International Conference on
  Learning Representations}, 2023. [Online]. Available:
  \url{https://openreview.net/forum?id=xE-LtsE-xx}
\BIBentrySTDinterwordspacing

\bibitem{suhail2022generalizable}
M.~Suhail, C.~Esteves, L.~Sigal, and A.~Makadia, ``Generalizable patch-based
  neural rendering,'' in \emph{Computer Vision--ECCV 2022: 17th European
  Conference, Tel Aviv, Israel, October 23--27, 2022, Proceedings, Part
  XXXII}.\hskip 1em plus 0.5em minus 0.4em\relax Springer, 2022, pp. 156--174.

\bibitem{andrew2001multiple}
A.~M. Andrew, ``Multiple view geometry in computer vision,'' \emph{Kybernetes},
  vol.~30, no. 9/10, pp. 1333--1341, 2001.

\bibitem{wang2004image}
Z.~Wang, A.~C. Bovik, H.~R. Sheikh, and E.~P. Simoncelli, ``Image quality
  assessment: from error visibility to structural similarity,'' \emph{IEEE
  transactions on image processing}, vol.~13, no.~4, pp. 600--612, 2004.

\bibitem{zhang2018unreasonable}
R.~Zhang, P.~Isola, A.~A. Efros, E.~Shechtman, and O.~Wang, ``The unreasonable
  effectiveness of deep features as a perceptual metric,'' in \emph{Proceedings
  of the IEEE conference on computer vision and pattern recognition}, 2018, pp.
  586--595.

\bibitem{chen2023improving}
S.~Chen, J.~Li, Y.~Zhang, and B.~Zou, ``Improving neural radiance fields with
  depth-aware optimization for novel view synthesis,'' \emph{arXiv preprint
  arXiv:2304.05218}, 2023.

\bibitem{zhou2023nerflix}
K.~Zhou, W.~Li, Y.~Wang, T.~Hu, N.~Jiang, X.~Han, and J.~Lu, ``Nerflix:
  High-quality neural view synthesis by learning a degradation-driven
  inter-viewpoint mixer,'' in \emph{Proceedings of the IEEE/CVF Conference on
  Computer Vision and Pattern Recognition}, 2023, pp. 12\,363--12\,374.

\bibitem{bao2019depth}
W.~Bao, W.-S. Lai, C.~Ma, X.~Zhang, Z.~Gao, and M.-H. Yang, ``Depth-aware video
  frame interpolation,'' in \emph{Proceedings of the IEEE/CVF conference on
  computer vision and pattern recognition}, 2019, pp. 3703--3712.

\bibitem{wang2021pwclo}
G.~Wang, X.~Wu, S.~Jiang, Z.~Liu, and H.~Wang, ``Efficient 3d deep lidar
  odometry,'' \emph{IEEE Transactions on Pattern Analysis and Machine
  Intelligence}, vol.~45, no.~5, pp. 5749--5765, 2023.

\bibitem{wang2020unsupervised}
G.~Wang, C.~Zhang, H.~Wang, J.~Wang, Y.~Wang, and X.~Wang, ``Unsupervised
  learning of depth, optical flow and pose with occlusion from 3d geometry,''
  \emph{IEEE Transactions on Intelligent Transportation Systems}, vol.~23,
  no.~1, pp. 308--320, 2020.

\bibitem{9466401}
Y.~Xu, X.~Xu, J.~Jiao, K.~Li, C.~Xu, and S.~He, ``Multi-view face synthesis via
  progressive face flow,'' \emph{IEEE Transactions on Image Processing},
  vol.~30, pp. 6024--6035, 2021.

\bibitem{9042874}
K.~Lv, H.~Sheng, Z.~Xiong, W.~Li, and L.~Zheng, ``Pose-based view synthesis for
  vehicles: A perspective aware method,'' \emph{IEEE Transactions on Image
  Processing}, vol.~29, pp. 5163--5174, 2020.

\bibitem{9320342}
A.~P.~S. Kohli, V.~Sitzmann, and G.~Wetzstein, ``Semantic implicit neural scene
  representations with semi-supervised training,'' in \emph{2020 International
  Conference on 3D Vision (3DV)}, 2020, pp. 423--433.

\end{thebibliography}

\begin{IEEEbiography}[{\includegraphics[width=1in,height=1.25in,clip,keepaspectratio]{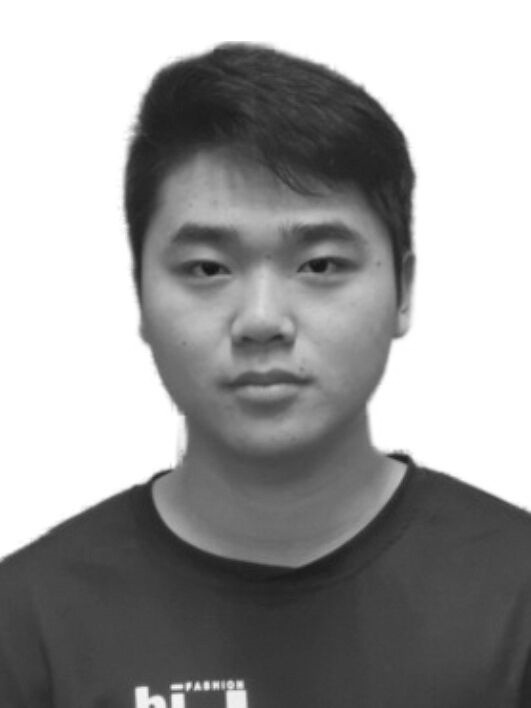}}]{Liang Song}
received the B.S. degree from Xuzhou University of Technology, Xuzhou, China, in 2021. He is currently pursuing the master degree in China University of Mining and Technology. His current research interests include computer vision and SLAM, in particular, Neural Radiance Field.
\end{IEEEbiography}

\begin{IEEEbiography}[{\includegraphics[width=1in,height=1.25in,clip,keepaspectratio]{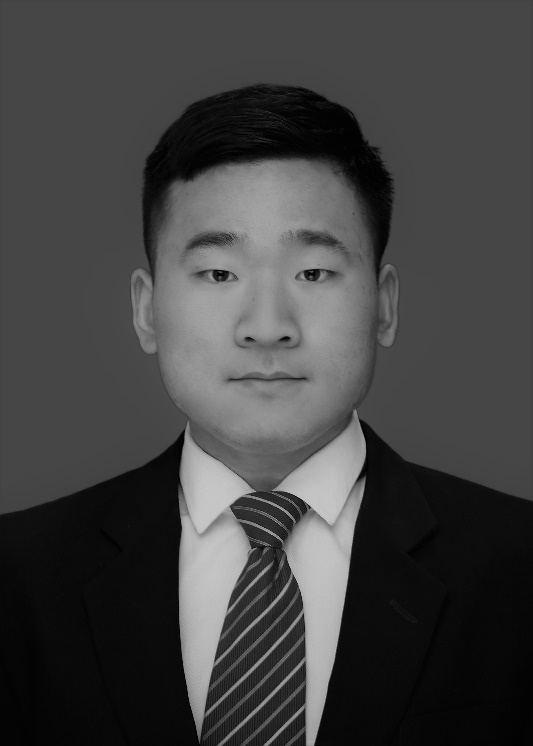}}]{Guangming Wang}
received the B.S. degree from Department of Automation from Central South University, Changsha, China, in 2018. He is currently pursuing the Ph.D. degree in Control Science and Engineering with Shanghai Jiao Tong University. His current research interests include SLAM and computer vision, in particular, neural radiance fields and neural rendering.
\end{IEEEbiography}

\begin{IEEEbiography}[{\includegraphics[width=1in,height=1.25in,clip,keepaspectratio]{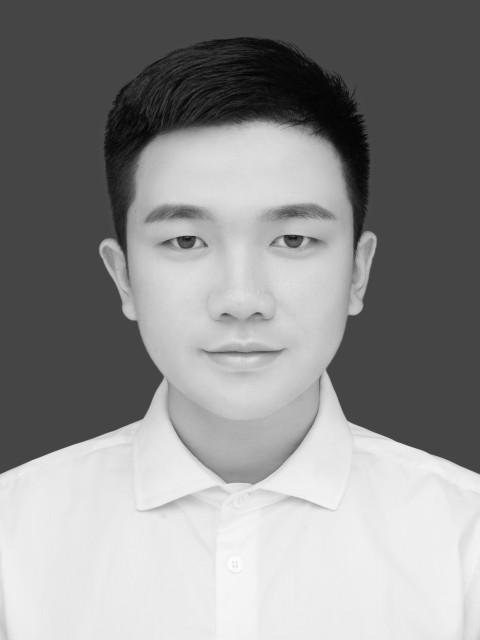}}]{Jiuming Liu}
received the B.S. degree from Department of Automation from Harbin Institute of Technology, Harbin, China, in 2022. He is currently pursuing the master degree in Control Science and Engineering with Shanghai Jiao Tong University. His current research interests include SLAM and computer vision, in particular, Neural Radiance Field, and transformer.
\end{IEEEbiography}

\begin{IEEEbiography}[{\includegraphics[width=1in,height=1.25in,clip,keepaspectratio]{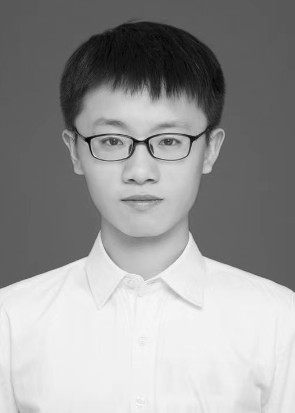}}]{Zhenyang Fu}
received the B.S. degree from Department of Automation from China University of Mining and Technology, Xuzhou, China, in 2023. His current research interests include intelligent visual perception and computer vision, in particular, 3D scene reconstruction and intelligent robots.
\end{IEEEbiography}

\begin{IEEEbiography}[{\includegraphics[width=1in,height=1.25in,clip,keepaspectratio]{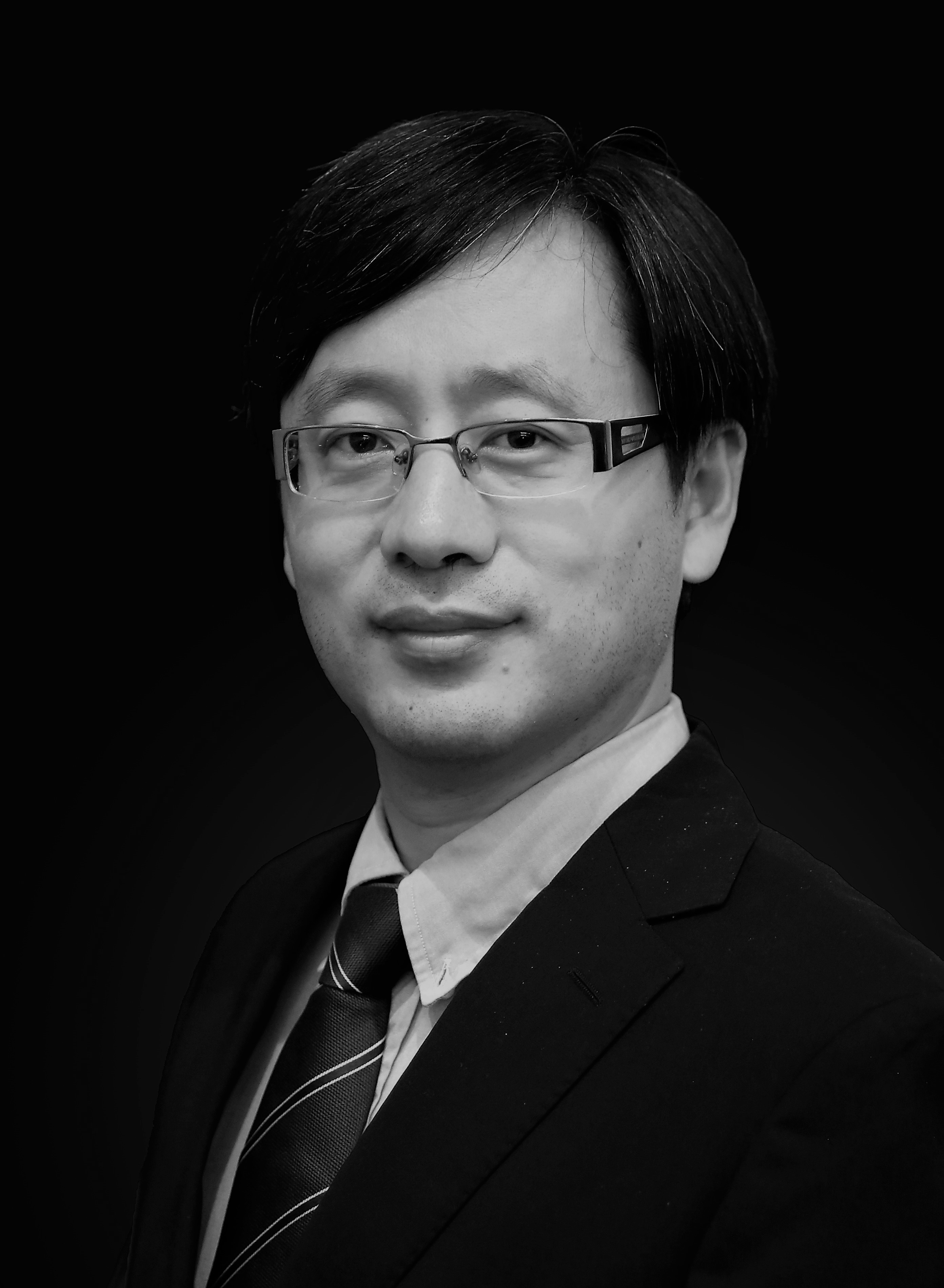}}]{Hesheng Wang}
received the B.Eng. degree in electrical engineering from the Harbin Institute of Technology, Harbin, China, in 2002, and the M.Phil. and Ph.D. degrees in automation and computer-aided engineering from The Chinese University of Hong Kong, Hong Kong, in 2004 and 2007, respectively. He is currently a Professor with the Department of Automation, Shanghai Jiao Tong University, Shanghai, China. His current research interests include visual servoing, service robot, computer vision, and autonomous driving. 
Dr. Wang is an Associate Editor of IEEE Transactions on Automation Science and Engineering, IEEE Robotics and Automation Letters, Assembly Automation and the International Journal of Humanoid Robotics, a Technical Editor of the IEEE/ASME Transactions on Mechatronics. He served as an Associate Editor of the IEEE Transactions on Robotics from 2015 to 2019. He was the General Chair of the IEEE RCAR 2016, and the Program Chair of the IEEE ROBIO 2014 and IEEE/ASME AIM 2019.
\end{IEEEbiography}

\begin{IEEEbiography}[{\includegraphics[width=1in,height=1.25in,clip,keepaspectratio]{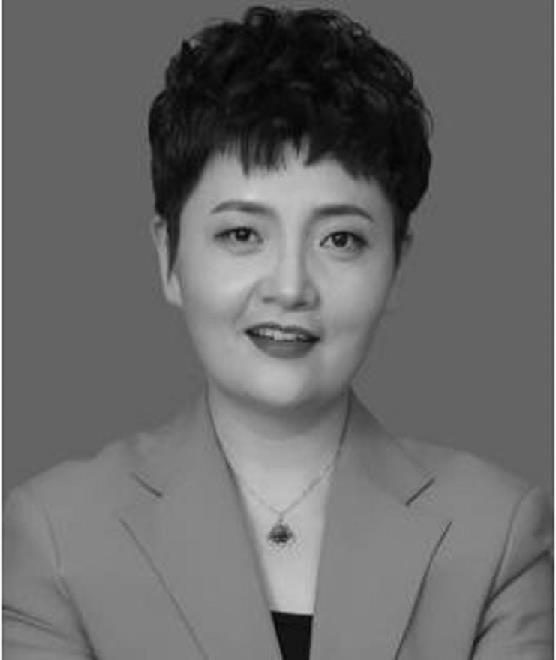}}]{Yanzi Miao}
received the Ph.D. degree in control science and engineering in 2009 from the China University of Mining and Technology, Xuzhou, China. As being a joint-PhD candidate and a visiting scholar, she worked in the Department of Informatics, University of Hamburg, Germany, in 2007 and 2017, respectively. Her is currently a Professor at the School of Information and control Engineering, China University of Mining and Technology. Her current research interests include Intelligent Perception and Fusion, Machine Vision and Active Olfaction. 
She has served as the Technical Co-Chair of IEEE RCAR 2019.
\end{IEEEbiography}




\end{document}